# SAMA-VTOL: A new unmanned aircraft system for remotely sensed data collection

Mohammad Reza Bayanlou [1] and Mehdi Khoshboresh-Masouleh [2]

1. Aerospace Engineering Department, Sharif University of Technology, Tehran, Iran.

   E-mail: mohammadreza.bayanlou@ae.sharif.edu, ORCID: https://orcid.org/0000-0003-4031-9238

2. School of Surveying and Geospatial Engineering, College of Engineering, University of Tehran, Tehran, Iran.

   E-mail: m.khoshboresh@ut.ac.ir, ORCID: https://orcid.org/0000-0002-9565-3615

**Abstract**

In recent years, unmanned aircraft systems (UASs) are frequently used in many different applications of photogrammetry such as building damage monitoring, archaeological mapping and vegetation monitoring. In this paper, a new state-of-the-art vertical take-off and landing fixed-wing UAS is proposed to robust photogrammetry missions, called SAMA-VTOL. In this study, the capability of SAMA-VTOL is investigated for generating orthophoto. The major stages are including designing, building and experimental scenario. First, a brief description of design and build is introduced. Next, an experiment was done to generate accurate orthophoto with minimum ground control points requirements. The processing step, which includes automatic aerial triangulation with camera calibration and model generation. In this regard, the Pix4Dmapper software was used to orientate the images, produce point clouds, creating digital surface model and generating orthophoto mosaic. Experimental results based on the test area covering 26.3 hectares indicate that our SAMA-VTOL performs well in the orthophoto mosaic task.

# 1. Introduction

In recent years, Unmanned Aircraft Systems (UASs) has been used in new applications such as marine wildlife detection [1], agricultural production management [2], moving object detection [3]–[6], search and rescue missions [7], disaster management [8], [9], real-time building damage mapping [10]–[13], 3D mapping [14]–[18], and automatic building monitoring [19], [20]. In this regard, one of the most comprehensive studies on the application of UASs is photogrammetry mission [8], [14], [21]–[23]. Generating orthophoto with UAS-based images is an important product for applied photogrammetry [24], [25]. Nowadays, the role of traditional methods such as terrestrial mapping and traditional aerial photogrammetry techniques, has been dimmed due to the high cost and also the need for a long time to produce an orthophoto [26]. An affordable and accurate way to generate orthophoto is to use the combination of UAS with a calibrated digital camera and the global positioning system [27]. Although this method also requires fieldwork, it requires less manpower than traditional methods.

Digital Elevation Model (DEM), Digital Surface Model (DSM) and orthophoto are useful geospatial products of photogrammetry missions. They are useful for all operators because they combine geometry and photorealism in order to provide a metric visualization of the real-world areas [28]. Particularly, they are essential datasets used in many different fields of geomatics science and geospatial engineering. Therefore, it is very important to provide an efficient solution for generating geospatial products with an easy and straightforward way [29]. The use of UAS is a good solution for generating geospatial products. The main purpose of this study was to design and construction of a new state-of-the-art vertical take-off and landing (VTOL) fixed-wing UAS for photogrammetry applications. The proposed UAS called SAMA-VTOL. The SAMA-VTOL introduces a variety of new features and capabilities for users. In this regard, the major contributions of this study to the remotely sensed data collection techniques are as follows:

(1) An original fixed-wing system is proposed for smart surveying.

(2) Flight safety has improved due to perform the VTOL.

(3) The use of an 18-megapixel digital camera to improvement the image quality.

(4) Integrate GNSS-PPK with SAMA-VTOL platform for reduces the time consuming field work.

These features make SAMA-VTOL well suited to photogrammetry applications. The proposed novel UAS, on paper, involves two main stages. First, conceptual design and construction process of the SAMA-VTOL. Second, performance evaluation of the SAMA-VTOL for generating DSM and orthophotos in case study research.

## 2. Methodology

### 2.1 SAMA-VTOL: A brief description of design and build

The proposed UAS based on conceptual designing to real prototyping workflow. The systematic conceptual layout flow diagram for designing the SAMA-VTOL is presented in Figure 1. The stages of systematic conceptual layout include mission requirement (e.g. flight time), platform configuration (e.g. stability and flight dynamics), mathematical modelling for UAS weight estimation, and wing configuration based on flight mission.

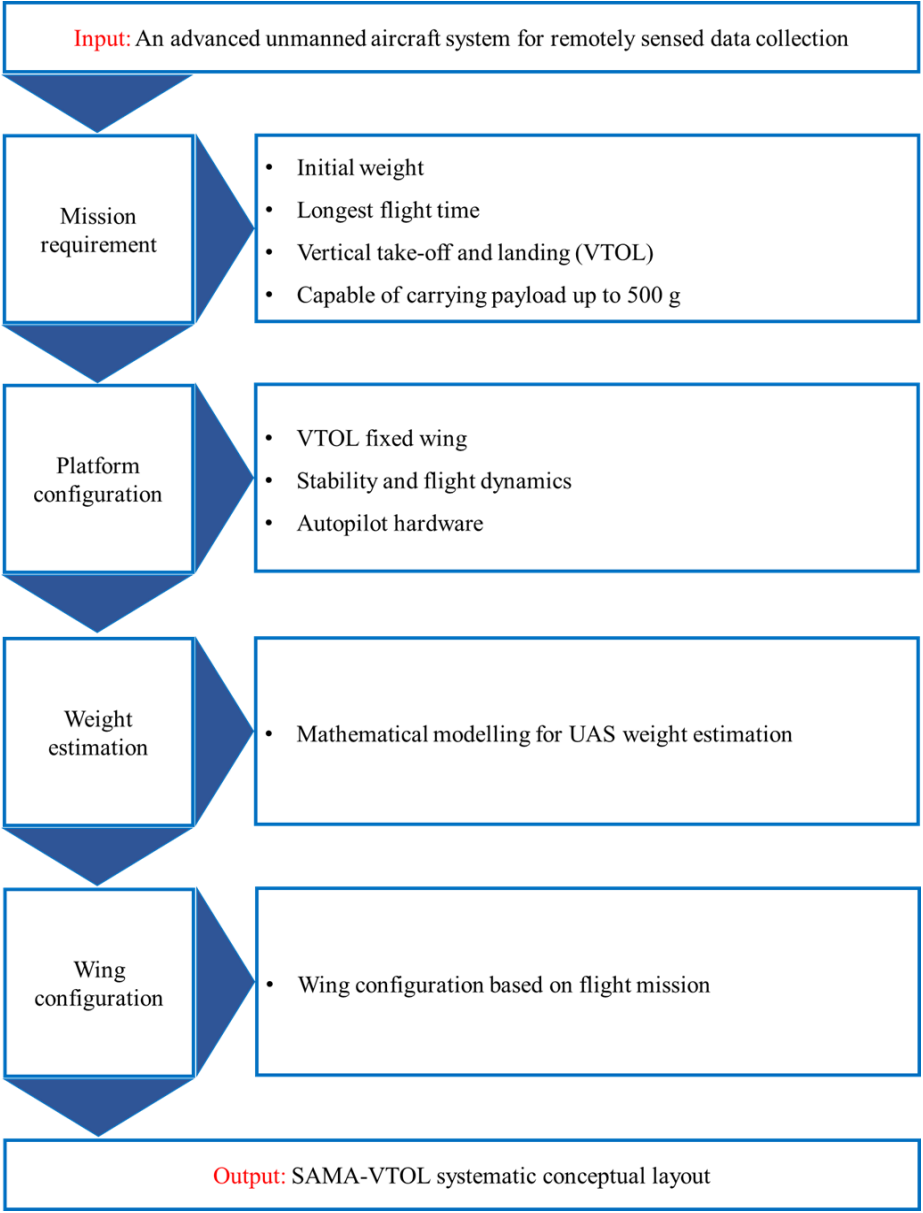

Figure 1. Systematic conceptual design flow diagram for SAMA-VTOL.

The main purpose of real prototyping stage is the iterative development (effective feedback for optimizing model), testing (checking activities), and refinement (highest performance) of the improved concept layout. The workflow from conceptual designing to real prototyping is presented in Figure 2. In this section, the SAMA-VTOL has been successfully built and tested for robust and safe flight.

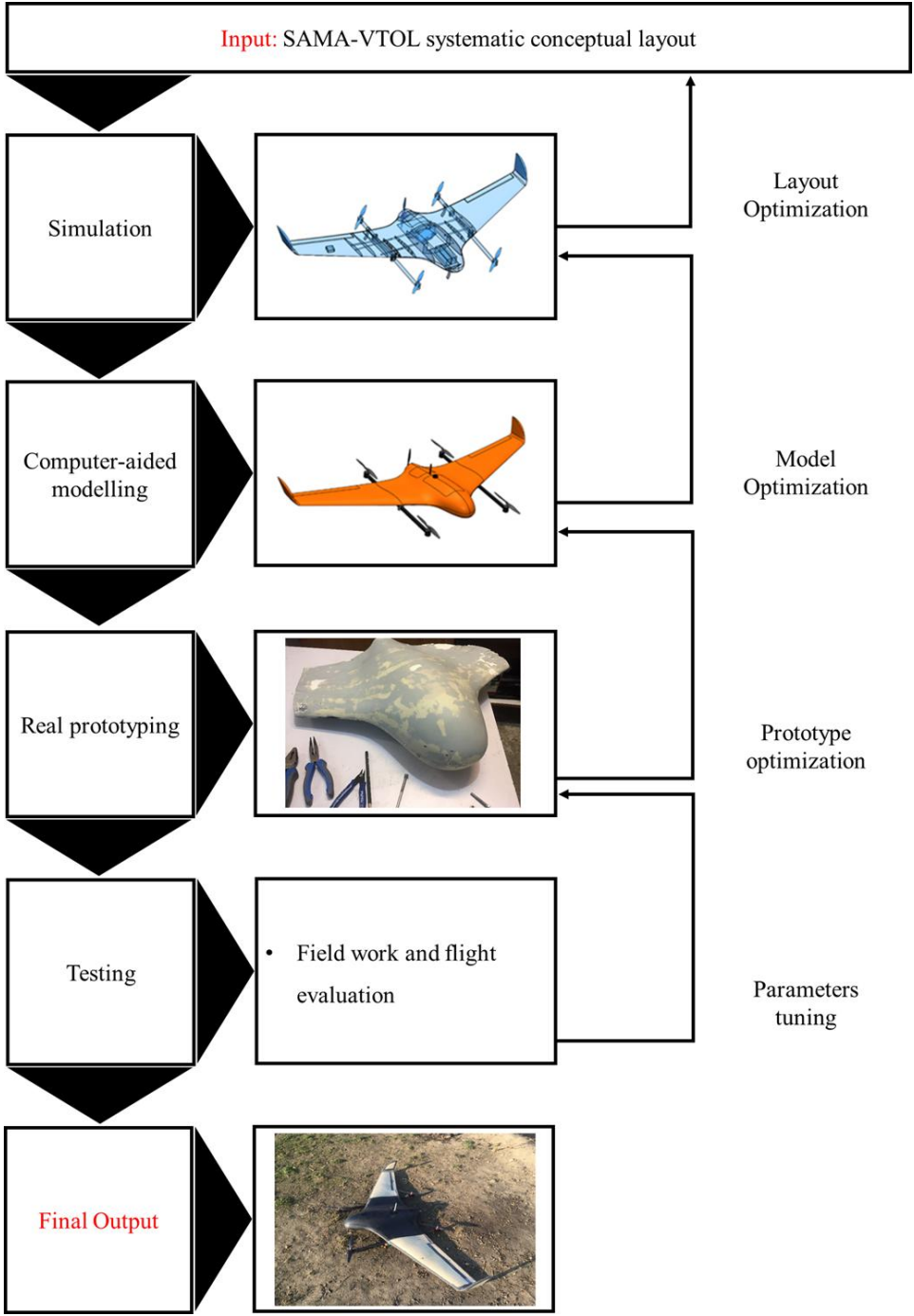

Figure 2. SAMA-VTOL design and manufacturing steps, (a) structure designing and sub-system placement, (b) final structure, (c) making the SAMA-VTOL body out of carbon fiber fabric, and (d) control parameters tuning.

## 2.2 Experimental scenario

As a metric for evaluating the SAMA-VTOL, we designed an experimental scenario. This scenario includes, the preparation of equipment to generate geospatial products (e.g. DSM and orthophoto). The workflow of the research is shown in Figure 3.

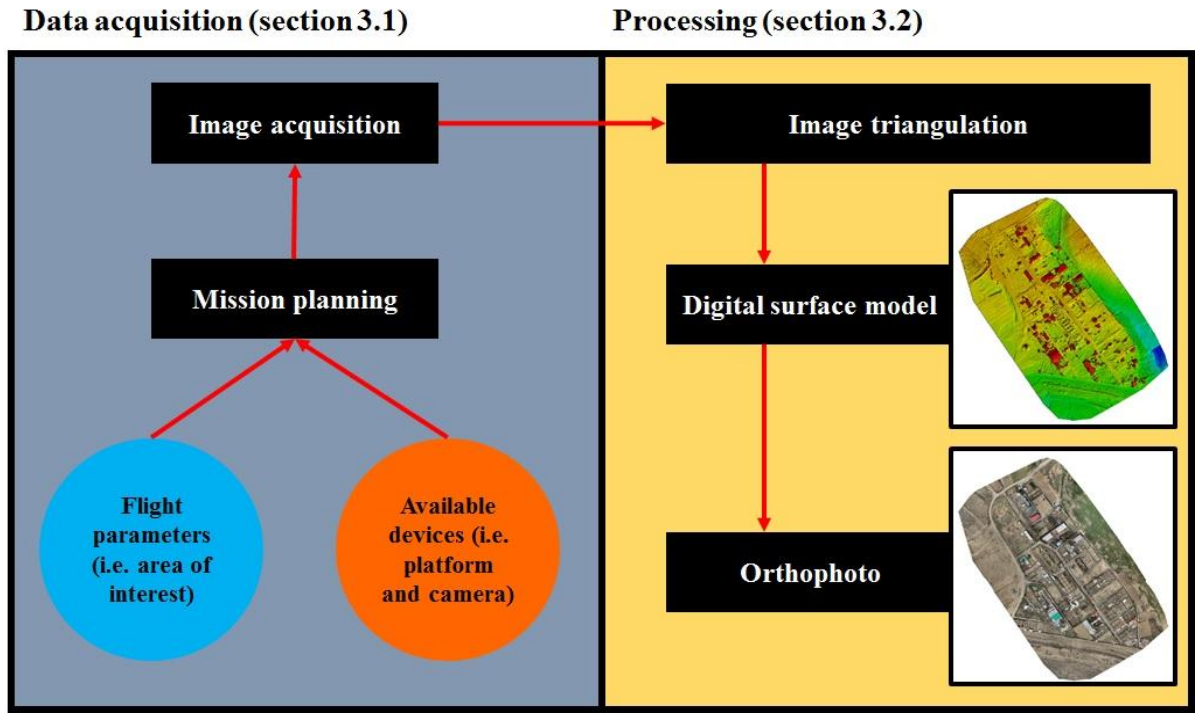

Figure 3. Workflow schematic as for evaluating the SAMA-VTOL.

## 3. Experiments

### 3.1 Data acquisition

The case study research is part of the Ahmadabad-e Mostowfi district, Iran. It covers a research area of 26.3 hectares inside the Ahmadabad-e Mostowfi district with an average altitude of 1090 m (Figure 4). The land cover consists of built-up areas and vegetation regions.

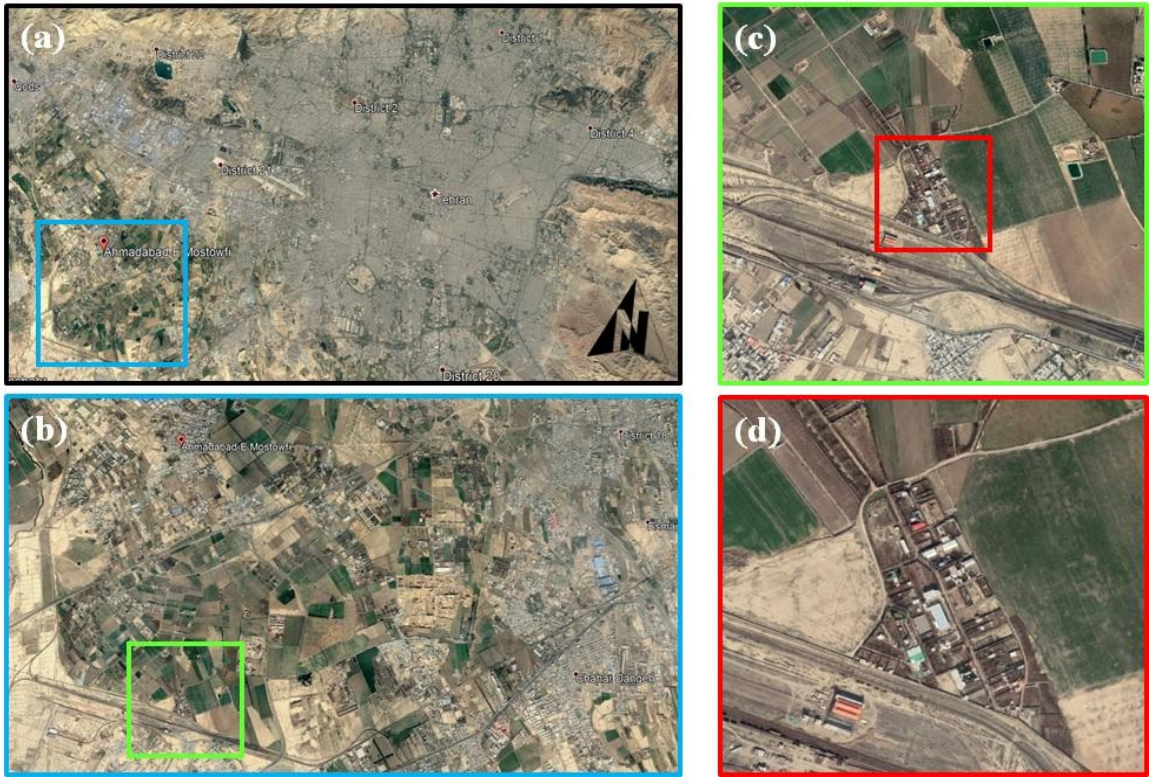

Figure 4. Overview of the research area and its location, (a) Tehran Province, (b) Ahmadabad-e Mostowfi district, (c) case study, and (d) test area.

In this study, SAMA-VTOL was equipped with a CANON EOS M digital camera (18M APS-C Sensor) to acquire georeferenced aerial images without ground control points. We established the checkpoints by using the Post-Processed Kinematic (PPK) positioning and 1×1 meter targets were made to mark various checkpoints throughout the case study (Figure 5). Moreover, the Pix4Dmapper software was used to orientate the images and produce point clouds, DSM and orthophoto mosaics and QGroundControl software was used to mission planning and flight control.

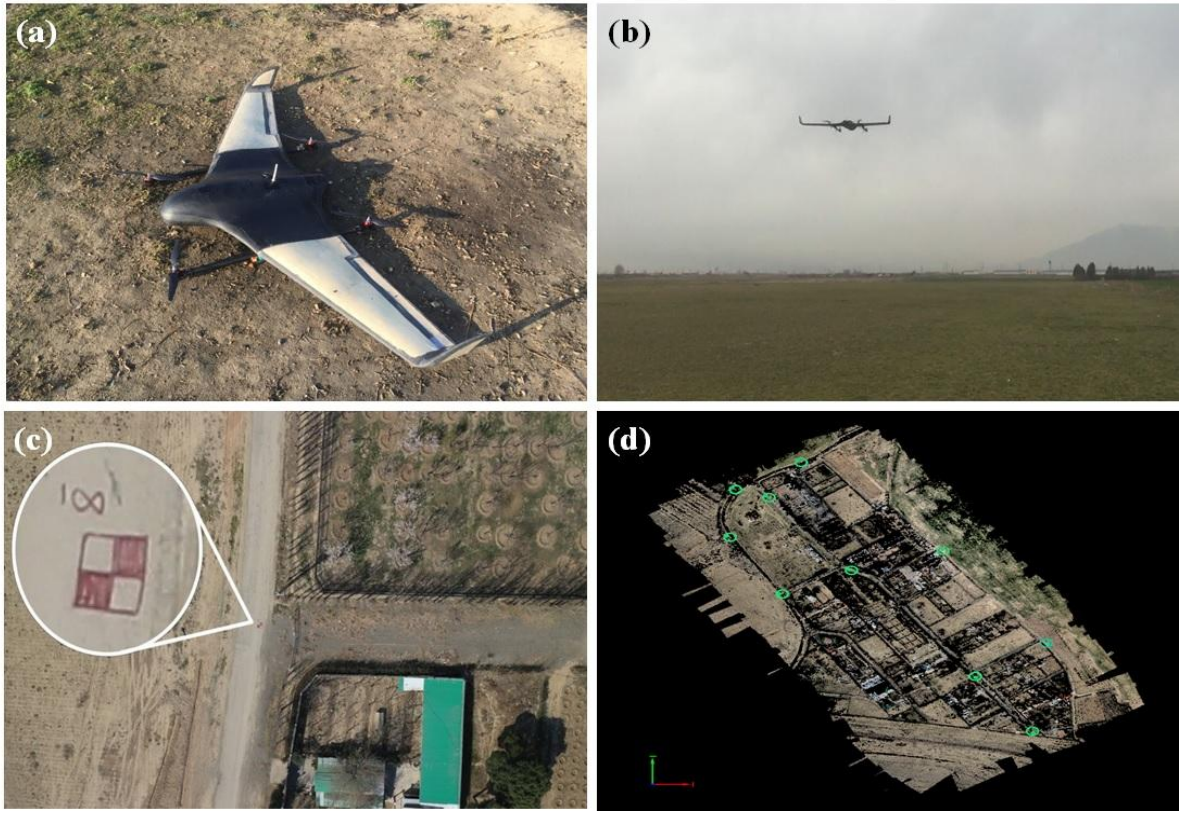

Figure 5. Overview of the research equipment, (a) SAMA-VTOL equipped with CANON EOS M digital camera, (b) SAMA-VTOL taking off, (c) checkpoint template, and (d) distribution of checkpoints.

**3.2 Processing**

The processing steps, includes aerial triangulation based bundle adjustment with camera calibration and model generation with Pix4Dmapper software. Pix4Dmapper using Scale-Invariant Feature Transform (SIFT) algorithm for accurate key points extraction (image matching) from single images [30]. In addition, the use of checkpoints helps to improving accuracy assessment stage in products generated. Table 1 shows the bundle block adjustment details, and Table 2 shows the summary initial and optimized results for camera calibration.

Table 1. Bundle block adjustment details.

| Number of 2-D keypoint | Number of 3-D points | Mean Reprojection Error |
|---|---|---|
| 5776111 | 2260338 | 0.202 pixels |

Table 2. Internal camera parameters.

|  | Focal Length | Principal Point | Radial lens distortions | | | | |
| --- | --- | --- | --- | --- | --- | --- | --- |
|  | - | - | K1 | K2 | K3 | T1 | T2 |
| **Initial Values** | 22.000 mm | 11.354 mm | 0.000 | 0.000 | 0.000 | 0.000 | 0.000 |
| **Optimized Values** | 21.949 mm | 11.547 mm | -0.011 | 0.085 | -0.151 | 0.000 | 0.000 |

### 3.3 Performance evaluation

Ahmadabad-e Mostowfi dataset contains 209 images that were acquired over a test area specifically prepared for this study. The flight height is 100 m while forward/side overlaps are 60% and 60%, respectively. Figure 6 shows the results of the DSM and orthophoto from the test region. Table 3 shows the error details of checkpoints in this test. Experimental results based on the test area indicate that our SAMA-VTOL performs well in the generating geospatial products task.

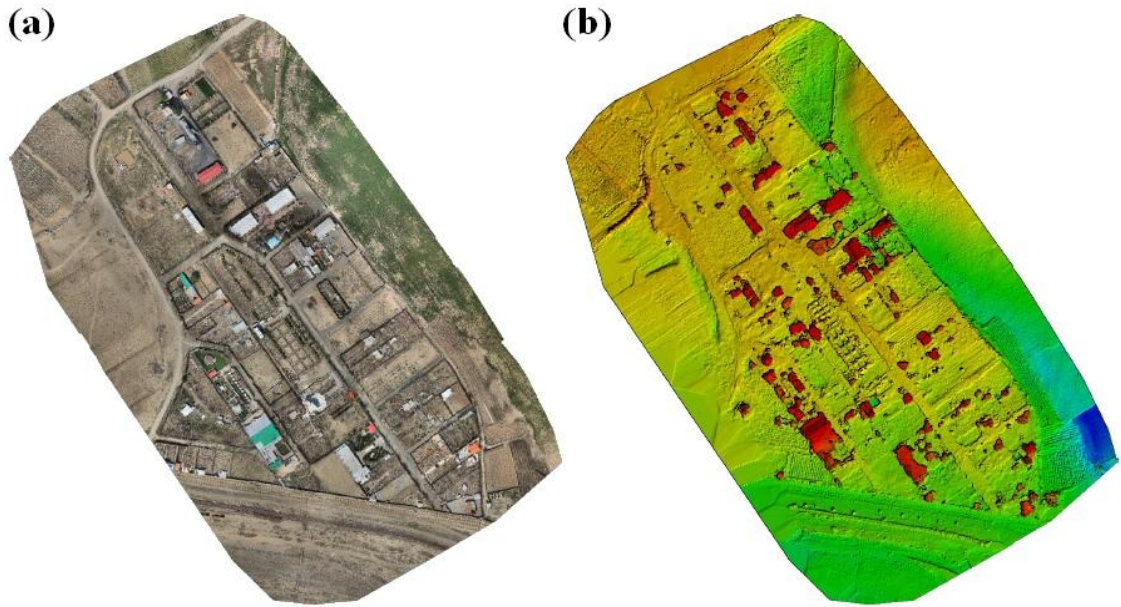

Figure 6. Products generated from Pix4Dmapper, (a) orthomosaic, and (b) digital surface model (DSM) with average ground sampling distance (GSD) equal to 1.91 cm.

Table 3. Error details of checkpoints.

| Error | Mean (meter) | Sigma (meter) | RMSE (meter) |
| --- | --- | --- | --- |
| X | -0.00326 | 0.03288 | 0.03304 |
| Y | -0.01404 | 0.03280 | 0.03855 |
| Z | 0.01792 | 0.03855 | 0.04251 |

## 4. Conclusions

In this paper, we proposed a new state-of-the-art VTOL fixed-wing UAS for photogrammetry missions, called SAMA-VTOL. The purpose of this study is to investigate the capabilities of the SAMA-VTOL in the field of photogrammetry, particularly, orthophoto generation. The results of the experiments in test area indicate that the SAMA-VTOL is affordable, and robust. Therefore, SAMA-VTOL is the accurate and advanced tool for professional surveying. In future work, we will expand and continue to improve SAMA-VTOL performance in larger areas.